\documentclass{article}
\usepackage{graphicx}
\usepackage{placeins}
\usepackage{svg}
\usepackage{subfloat}
\usepackage{todonotes}
\usepackage{flafter}
\usepackage{algpseudocode,algorithm,algorithmicx}
\usepackage{framed}

\newcommand{\figwidth}{0.9\linewidth}

\newcommand*\focusstart{\color{ForestGreen}}
\newcommand*\focusend{\color{black}}
\newcommand*\Let[2]{\State #1 $\gets$ #2}

\algrenewcomment[1]{\focusstart \hspace{1em} \(//\) #1 \focusend}
\algnewcommand{\LineComment}[1]{\State \(//\) #1}

\algnewcommand\algorithmicforeach{\textbf{for each}}
\algdef{S}[FOR]{ForEach}[1]{\algorithmicforeach\ #1\ \algorithmicdo}

\title{Finding shorter paths for robot arms using their redundancy}
\author{Scott Paulin, Tom Botterill, XiaoQi Chen, Richard Green}
\newcommand*\mathsub[1]{_{\textrm{\tiny #1\normalsize}}}
\newcommand{\exptwidth}{0.3}

\newcommand{\citet}[1]{\citeauthor{#1} \citeshell{\citeyear{#1}}}

\begin{document}
\maketitle

\begin{abstract}
Many robot arms can accomplish one task using many different joint configurations. Often only one of these configurations is used as a goal by the path planner. Ideally the robot's path planner would be able to use the extra configurations to find higher quality paths. In this paper we use the extra goal configurations to find significantly shorter paths that are faster to execute compared to a planner that chooses one goal configuration arbitrarily. In a grape vine pruning robot arm experiment our proposed approach reduced execution times by 58\%.
\end{abstract}

\section{Introduction}

Robots are now being used in uncontrolled environments, e.g. agriculture~\cite{Bac2014}, and must compute path plans online as obstacles are detected or new goals are identified. These robots may use a path planner to find collision free paths for the robot to follow. Ideally the robot's path planner will find paths that are quick to execute so that the robot can have a low cycle time. 

\begin{figure}
	\centering
	\subfloat{\includegraphics[width=\exptwidth\linewidth]{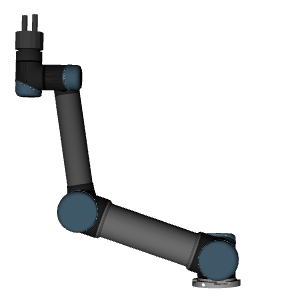}} \hspace{5mm}
	\subfloat{\includegraphics[width=\exptwidth\linewidth]{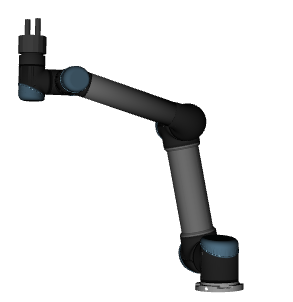}}
	\caption{Two poses for the UR5 robot arm that put the end-effector at the same position.}
	\label{fig:two_poses_same_ee_pos}
\end{figure}

Robot arm tasks are specified in their workspace for many applications, e.g. move the end-effector to a specific Cartesian position. These tasks can often be accomplished by the robot arm in many different poses (Fig.~\ref{fig:two_poses_same_ee_pos}), where a pose is defined as the workspace position of every part of the robot arm (not just the end-effector). We will define the different poses that can be used to achieve a task to be the robot arm's \emph{workspace redundancy}.

Some robot arms have joints that are capable of making more than one full revolution, e.g. Universal Robot's (UR) popular UR3, UR5, UR10 and ABB's IRB 2400 robot arms. These robot arms have \emph{configuration space redundancy} where each robot pose can be represented by a number of different configurations. For example, each of the UR5 arm poses shown in Fig.~\ref{fig:two_poses_same_ee_pos} can be achieved with 64 different configurations (see Sec.~\ref{sec:config_space_redundancy}). We will call these configurations that lead to the same robot pose \emph{equivalent configurations}. In this paper we investigate the effect of using the robot arm's workspace and configuration space redundancy in path planning to find shorter paths using an asymptotically optimal planner. 

\section{Background to improving planner performance using workspace redundancy}
Path planning is often performed in the robot's configuration space. In many cases only one~\cite{Coleman2015, Hirano2005, Lee2014a, Stilman2007} of the many possible workspace poses are considered by the path planner. 

One approach to exploiting a robot's redundancy is to allow the path planner to use multiple inverse kinematic solutions as a goal. This approach has been used with feasible path planners to improve their success rates~\cite{Dalibard2009, Ellekilde2013}. This approach is simple, but requires a fixed number of inverse kinematic solutions to be computed before planning begins. The robot's workspace redundancy can be exploited by using a large number of inverse kinematic solutions. Computing these solutions can be computationally expensive, and a large number of solutions may be unnecessary for some planning queries. Computing too few inverse kinematic solutions may lower the success rate of the planner.

Inverse kinematic solutions can be instead computed during planning from a \emph{workspace goal region}~\cite{Berenson2009,Berenson2011}. This region encapsulates the workspace poses that the robot can use to achieve its task. Inverse kinematic solutions satisfying the workspace goal region are computed and added to the planner's goal representation during planning. An advantage of this approach is that more goal configurations are considered on difficult planning queries, when the planner takes a long time to find a solution. Fewer inverse kinematic solutions are computed on simpler queries when the planner quickly finds a path. 

Some planners are capable of finding paths to goals in the robot's workspace without using an inverse kinematics routine~\cite{VandeWeghe2007}. The planner's exploration of configuration space is guided by a local controller that uses the transpose of the robot's Jacobian to minimise the distance between explored configurations and the robot's goal specified in the workspace. A number of other approaches use problem-specific heuristics~\cite{Bertram2006,Drumwright2006,Keselman2014,Stollenga2013} to guide the planner's search.

The planning approaches covered so far were developed to reduce computation time and improve the success rates of feasible path planners. Many of them could be extended for use with asymptotically optimal path planners, but there has been little work investigating the effects of using workspace redundancy with asymptotically optimal planners.

Dragan et~al~\cite{Dragan2011} modified the Covariant Hamiltonian Optimization for Motion Planning (CHOMP)~\cite{Zucker2013} trajectory optimizer to be able to handle a set of goal configurations. They found that considering a set of goals improved the paths that were found by CHOMP.  These results suggest that specifying a goal as a set of configurations may allow asymptotically optimal planners, e.g. RRTConnect*~\cite{Akgun2011,Jordan2013, Klemm2015}, to find better paths, which is what we investigate in this paper. 

Dragan~et~al~\cite{Dragan_2011_6887} attempt to predict goal configurations that would lead to high quality paths being found with CHOMP using a range of machine learning algorithms. Given a set of goal configurations, they manage to select one that allows the planner to find solutions that are on average 8\% worse than those found using the entire goal set. In this case it is better to use CHOMP with the entire goal set because the set of goal configurations has already been computed.

\section{Configuration space redundancy}
\label{sec:config_space_redundancy}
\begin{figure}
	\centering
	\includesvg[svgpath=images/, width=0.5\linewidth]{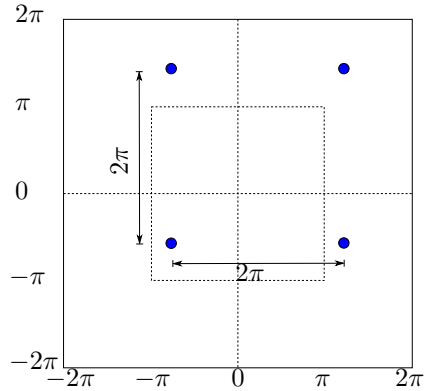}
	\caption{Equivalent configurations shown in blue for a robot arm with two joints that can each operate in the range $[-2\pi, 2\pi)$.}
	\label{fig:equivalent_configurations}
\end{figure}

Some robot arms, such as the UR5, have joints that can perform more than one full rotation. Given one configuration we can construct others that put all of the robot’s links in exactly the same positions by rotating any of the joints by one or more full revolutions. We call these \emph{equivalent configurations}, because they put each part of the robot in the same position but are distinct in the robot's configuration space. The configuration space of a two degree of freedom robot with equivalent configurations is shown in Fig.~\ref{fig:equivalent_configurations}.

Equivalent configurations are always separated by multiples $2\pi$ in each dimension that represents a rotational joint in the robot's configuration space. They can be calculated by adding or subtracting multiples of $2\pi$ from positions of rotational joints in the robot's configuration space, while remaining within the joint limits.

The number of equivalent configurations a robot arm has depends on the number of rotations, $n$, each of its $N$ joints can make:
\begin{equation}\label{eq:num_equivalent_configurations}
n\mathsub{equivalent} = \displaystyle\prod_{n=1}^{N} n_i 
\end{equation}

A robot arm that can make two full rotations with each of its two joints will have four equivalent configurations as shown in Fig.~\ref{fig:equivalent_configurations}. A robot arm with joints that can each only make one one full rotation will only have one equivalent configuration. This can also be seen in Fig.~\ref{fig:equivalent_configurations} where there is only one configuration in each $2\pi$ by $2\pi$ block.

\FloatBarrier
\section{Finding shorter paths using configuration space and workspace redundancy}

The robot's workspace and configuration space redundancy are used to compute a set of goal configurations as shown in Fig.~\ref{alg:get_goal_configurations}. These configurations are then used to represent the planner's goal. An inverse kinematic solver is used to generate a predefined number of configurations that satisfy the workspace goal and put the robot arm in distinct poses. These configurations represent the robot's workspace redundancy with respect to the task. The equivalent configurations for each of these inverse kinematics solutions is then computed. The resulting configurations represent the robot's workspace and configuration space redundancy with respect to the task. These configurations are then used to represent the path planner's goal.

\begin{figure}[htbp]
	\begin{framed}
		\begin{algorithmic}[1]
			\Function{ComputeGoalConfigurations}{$task\textunderscore goal$}
			\Let{goal\textunderscore configs}{\{\}}
			\Let{ws\textunderscore goal\textunderscore configs}{Inverse solutions that satisfy task\textunderscore goal}
			\ForEach {g $\in$  ws\textunderscore goal\textunderscore configs}
			    \Let{equiv\textunderscore configs}{Equivalent configurations of g}
			    \State Append equiv\textunderscore configs to goal\textunderscore configs
			\EndFor
			\State \Return goal\textunderscore configs
			\EndFunction
		\end{algorithmic}
	\end{framed}
	\caption{Algorithm for computing goal configurations for a robot using its workspace and configuration space redundancy.}\label{alg:get_goal_configurations}
\end{figure}

\section{Experiments}

\begin{figure}[htb]
	\centering
	\subfloat[Vine to be pruned.]{\includegraphics[height=\exptwidth\linewidth]{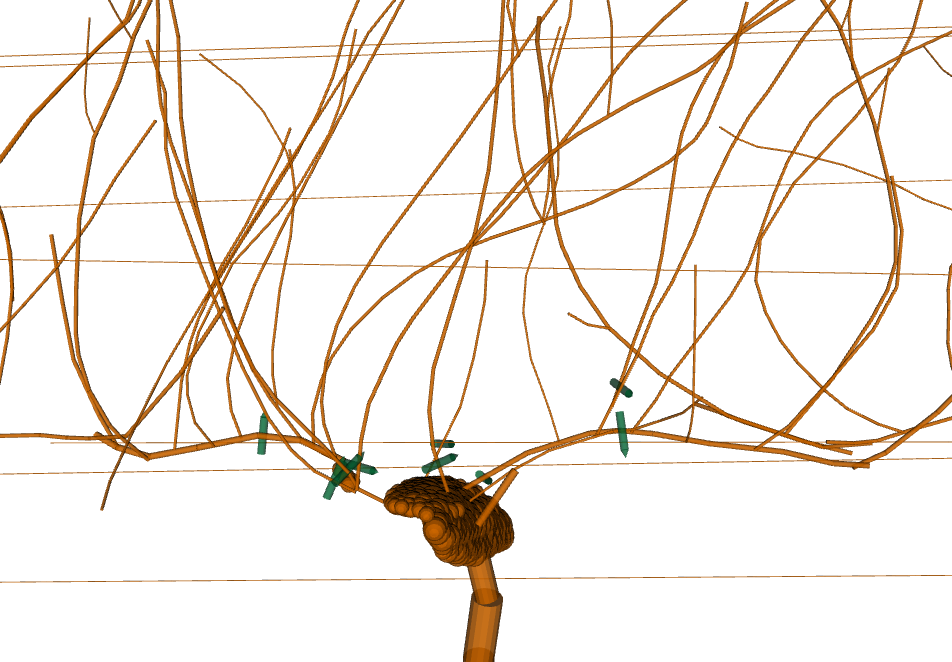}}\hspace{5em}
	\subfloat[The robot arm in a cutting position.]{\includegraphics[height=\exptwidth\linewidth]{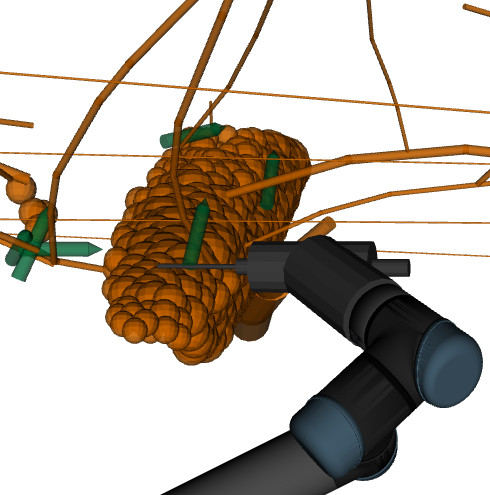}}
	\caption{Vine pruning scenario. Cutpoints are shown in green.}
	\label{fig:vine_pruning_robot}
\end{figure}

\begin{figure}[htb]
	\centering
	\subfloat[Robot arm with gripper model.]{\includegraphics[width=\exptwidth\linewidth]{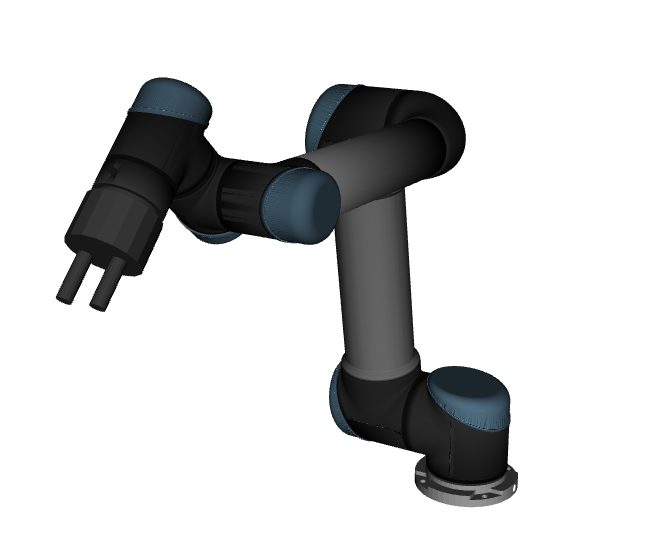}}\hspace{5em}
	\subfloat[Robot arm reaching into a cubicle.]{\includegraphics[width=\exptwidth\linewidth]{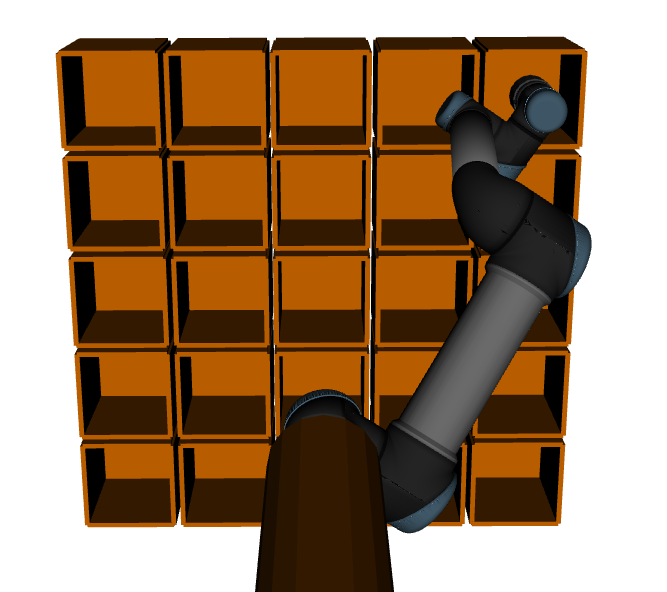}}
	\caption{Cubicle picking scenario.}
	\label{fig:cubicle_picking_robot}
\end{figure}

We test the impact using multiple goals has on the performance of RRTConnect* with integrated short-cutting planner~\cite{Paulin2016a} configured to minimise the Euclidean path length. For both experiments the Euclidean length (sum of Euclidean lengths of each path segment, in radians) and execution time (how long it would take the robot arm to follow the path) were recorded. We tested the planner's performance when it was used with a different numbers of the closest goal configurations, and when it was used with different numbers of random goal configurations. Both cases were tested on two robots, one for pruning grape vines~\cite{BotterillVPR} (Fig.~\ref{fig:vine_pruning_robot}) and another for reaching into cubicles (Fig.~\ref{fig:cubicle_picking_robot}).

On the vine pruning robot, the planners were tasked with moving the robot arm to cut positions on the vine. This task involves planning fine motions around thin obstacles. To speed up planning we used a collision detector that was specialised for use in this problem~\cite{Paulin2015b}. We limited the cuts used to those where five robot goal poses could be sampled. The 16 equivalent configurations for each of these were calculated and used as goal configurations, making a maximum of 80 goals per cut. Each robot pose had 16 equivalent configurations by equation~\ref{eq:num_equivalent_configurations} because two of the UR5's six joints had to be limited for this task (see Appendix~\ref{app:joint_limiting}).

The cubicle picking environment was designed to be similar to that used in previous research~\cite{POMPChoudhury2016, Phillips2012, Ratliff2009} and the 2015 Amazon Picking Challenge~\cite{Correll2016}. The planner had to compute plans so that the robot arm would reach from its start position in one cubicle into another. Exiting the start cubicle and entering the goal cubicle both required fine motion plans. We used the Flexible Collision Library (FCL)~\cite{Pan2012} for collision detection. An analytical inverse kinematics solver for the UR5~\cite{Hawkins2013} was used to generate the eight robot arm poses to reach the arm into the centre of each cubicle with a fixed end-effector orientation. The 32 equivalent configurations for each of these poses were calculated and used as goal configurations, meaning there were 256 goal configurations per cubicle. Each robot pose had 32 equivalent configurations by equation~\ref{eq:num_equivalent_configurations} because one of the joints had to be limited for this task (see Appendix~\ref{app:joint_limiting}).

\section{Results}

\begin{figure}
\centering
\includegraphics[width=\figwidth]{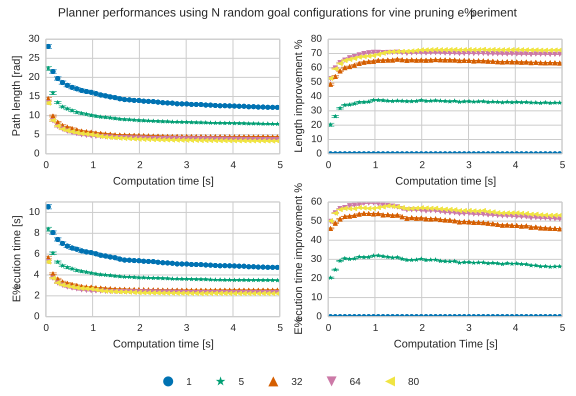}
\caption{Path length and execution times using different numbers of random goals (left) and the improvement over using one goal (right) for the vine pruning experiment. Error bars show the 95\% confidence interval.}
\label{fig:vpr_random_all_eq_len_exec}
\end{figure}

\begin{figure}
\centering
\includegraphics[width=\figwidth]{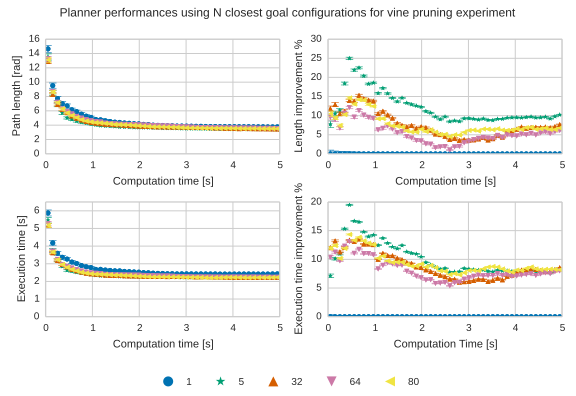}
\caption{Path length and execution times using different numbers of the closest goal configurations (left) and the improvement over using one goal (right) for the vine pruning experiment. Error bars show the 95\% confidence interval.}
\label{fig:vpr_sorted_all_eq_len_exec}
\end{figure}

\begin{figure}
	\centering
	\includegraphics[width=\figwidth]{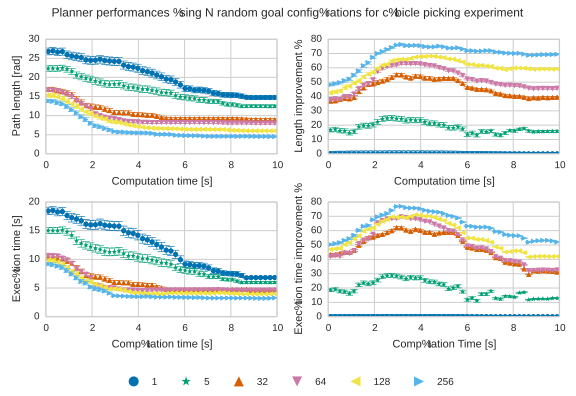}
	\caption{Path length and execution times using different numbers of random goals (left) and the improvement over using one goal (right) for the cubicles experiment. Error bars show the 95\% confidence interval.}
	\label{fig:cub_random_all_eq_len_exec}
\end{figure}

\begin{figure}
	\centering
	\includegraphics[width=\figwidth]{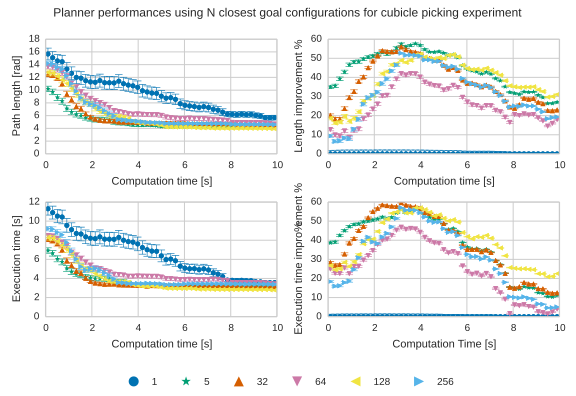}
	\caption{Path length and execution times using different numbers of closest goal configurations (left) and the improvement over using one goal (right) for the cubicles experiment. Error bars show the 95\% confidence interval.}
	\label{fig:cub_sorted_all_eq_len_exec}
\end{figure}

We performed two experiments on the vine pruning and cubicle picking robots. In the first we tested the planner's performance with different numbers of randomly-chosen goal configurations for each target (cubicle or cut). This is to simulate the case where a randomised inverse kinematics solver is being used and the user does not know what the closest goal configurations are, these results are shown in Figs.~\ref{fig:vpr_random_all_eq_len_exec},~\ref{fig:cub_random_all_eq_len_exec}. In the second experiment we tested the planner's performance with different numbers of the closest goal configurations. This is to determine the influence that the proximity of the goal configuration to the start configuration has, these results are shown in Figs.~\ref{fig:vpr_sorted_all_eq_len_exec},~\ref{fig:cub_sorted_all_eq_len_exec}. In both experiments we tested planning to 303 cuts for the vine pruning robot and into 100 cubicles for the cubicle picking robot. 

\begin{figure}
	\centering
	\includegraphics[width=\figwidth]{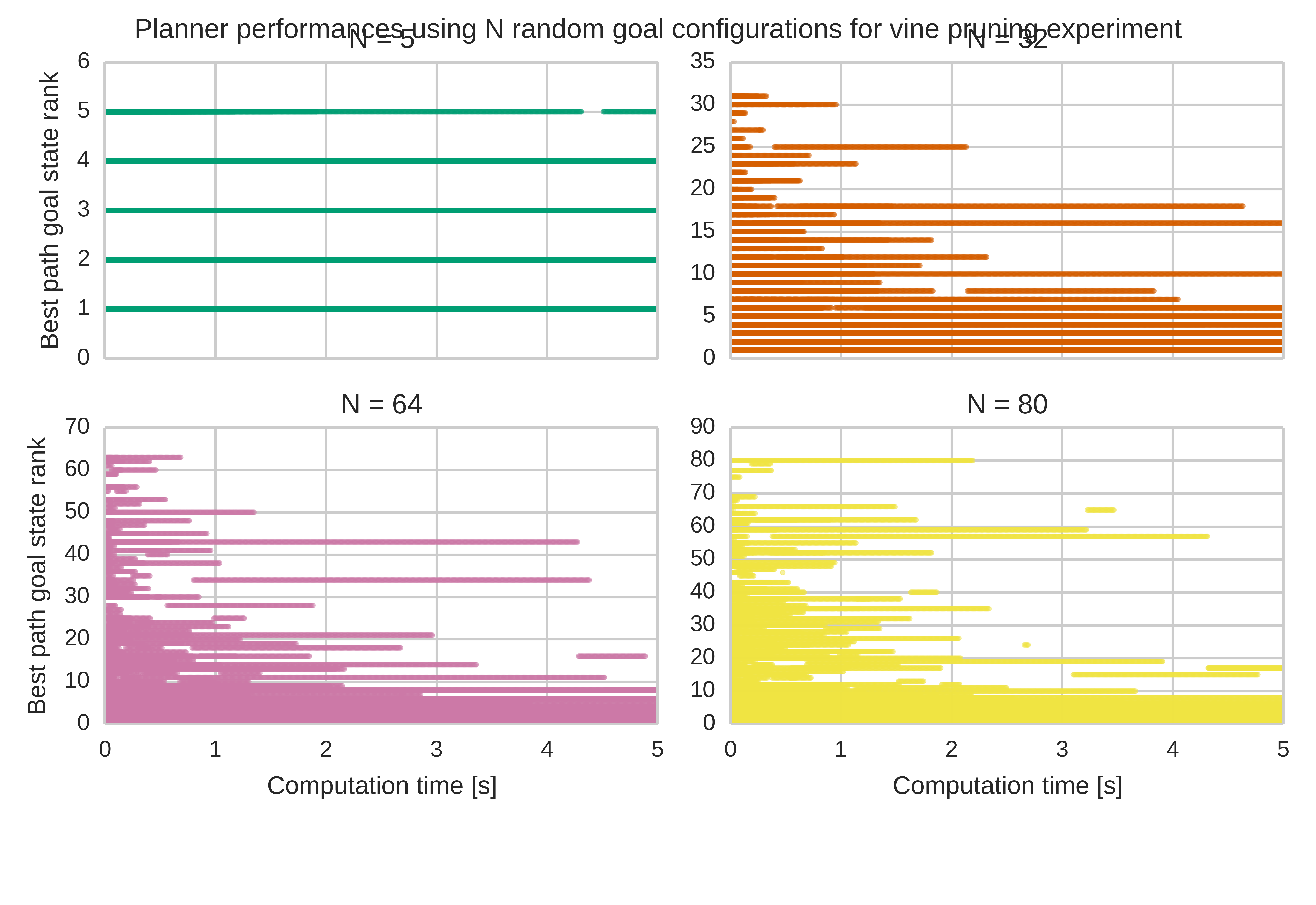}
	\caption{Ranking of goal configuration used in shortest path over time for vine pruning experiment. The shortest path found by the planner is frequently not to the closest in configuration space.}
	\label{fig:vpr_random_all_eq_eq_used}
\end{figure}

\begin{figure}
	\centering
	\includegraphics[width=\figwidth]{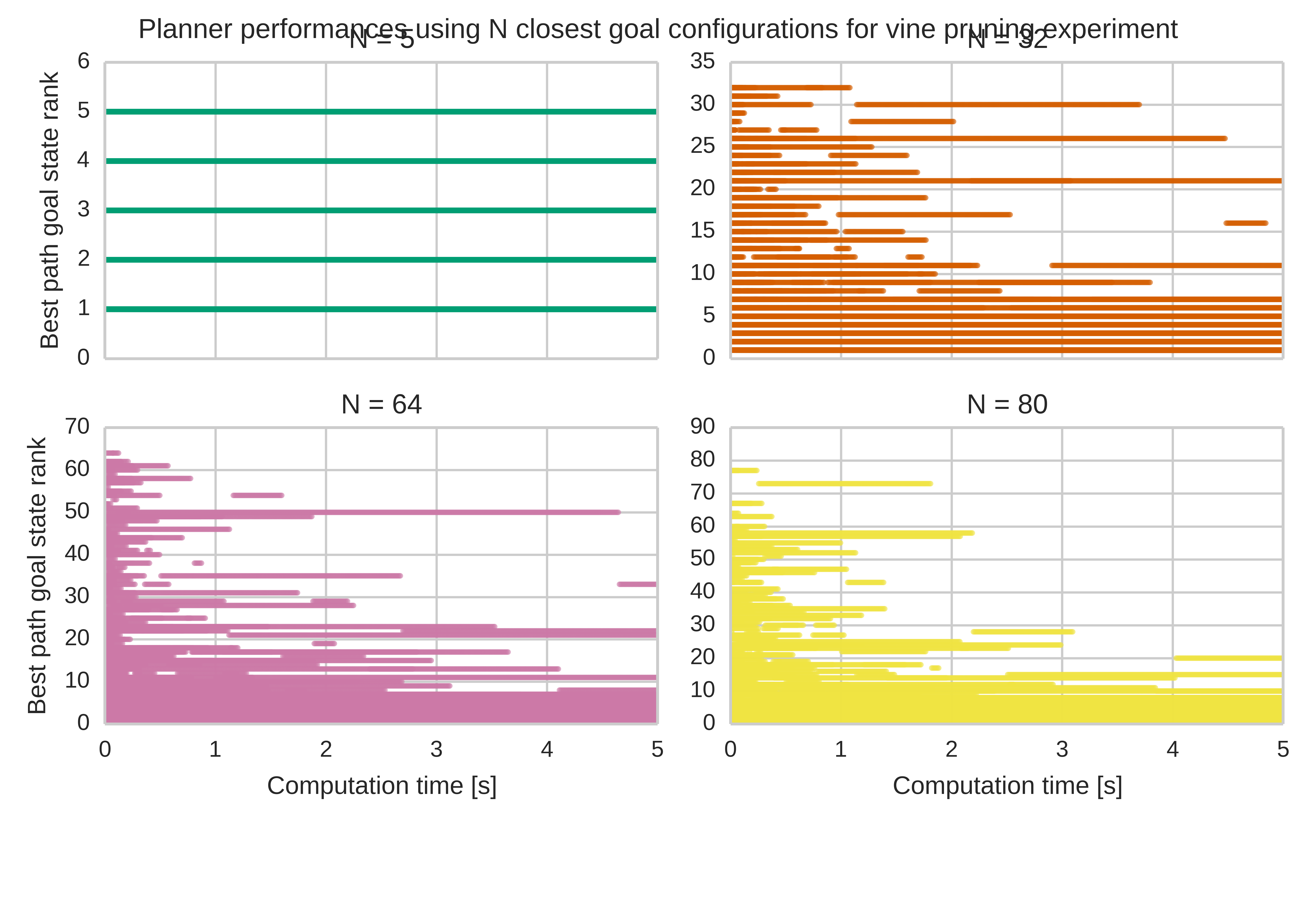}
	\caption{Ranking of goal configuration used in shortest path over time for vine pruning experiment. The shortest path found by the planner is frequently not to the closest in configuration space.}
	\label{fig:vpr_sorted_all_eq_eq_used}
\end{figure}

\begin{figure}
	\centering
	\includegraphics[width=\figwidth]{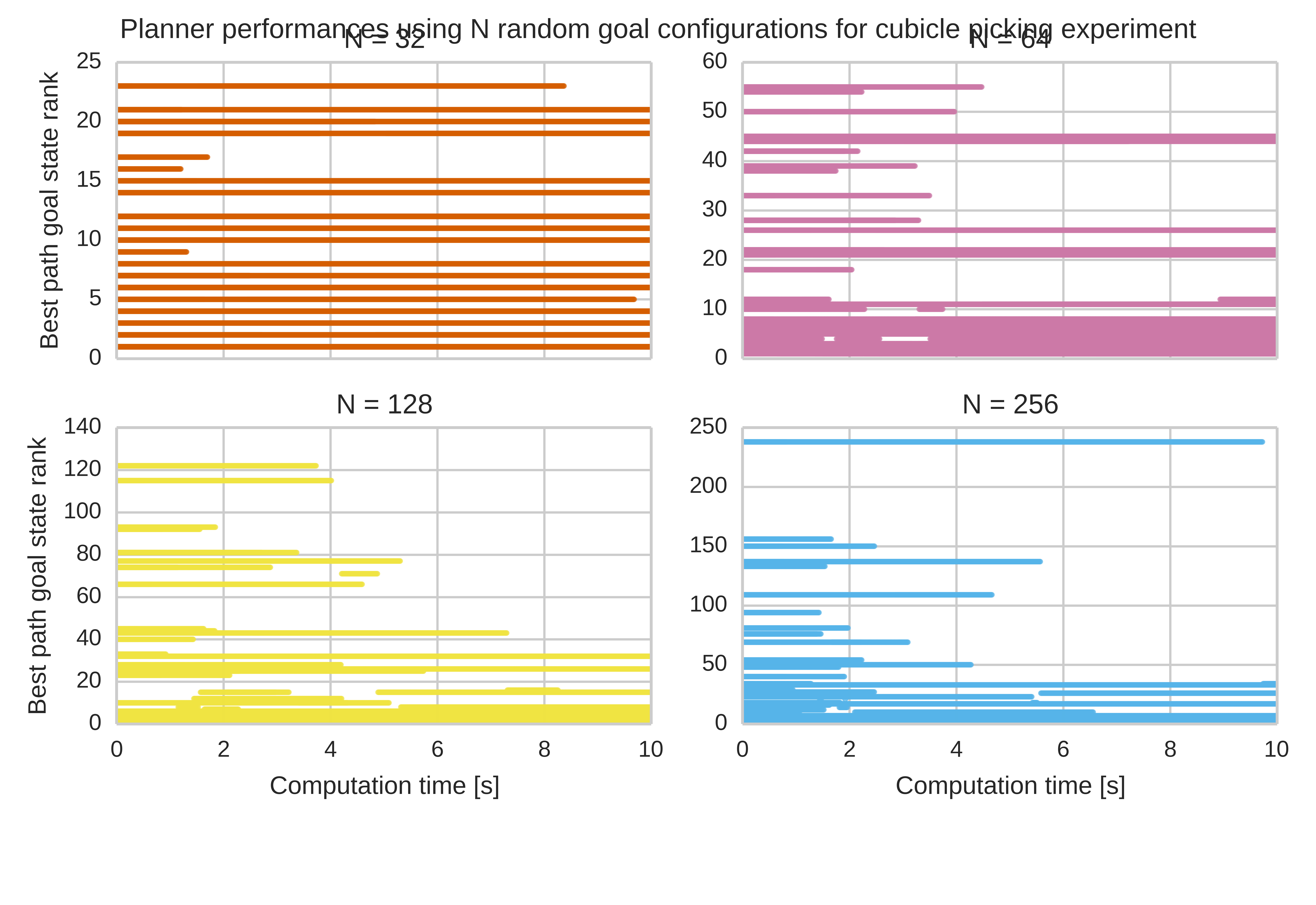}
	\caption{Ranking of goal configuration used in shortest path over time for cubicles experiment. The shortest path found by the planner is frequently not to the closest in configuration space.}
	\label{fig:cub_random_all_eq_eq_used}
\end{figure}

\begin{figure}
	\centering
	\includegraphics[width=\figwidth]{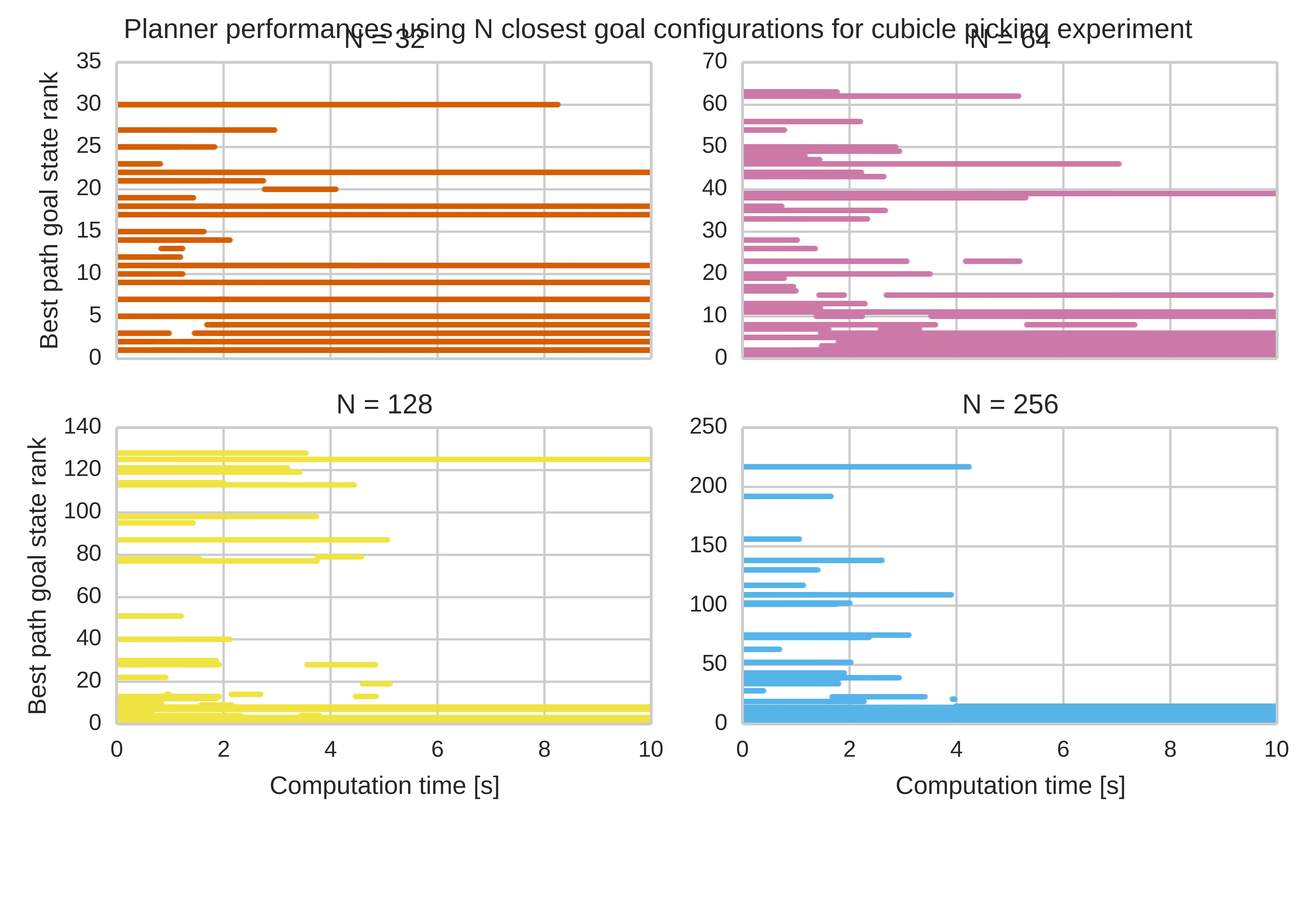}
	\caption{Ranking of goal configuration used in shortest path over time for cubicles experiment. The shortest path found by the planner is frequently not to the closest in configuration space.}
	\label{fig:cub_sorted_all_eq_eq_used}
\end{figure}

The ranking for the goal configuration used in the shortest path was stored during planning. A ranking of $n$ means that the $n$\textsuperscript{th} closest goal configuration to the start is being used in the shortest path. The goal configuration rankings for the vine pruning and cubicle picking experiments are shown in Figs.~\ref{fig:vpr_random_all_eq_eq_used},~\ref{fig:vpr_sorted_all_eq_eq_used},~\ref{fig:cub_random_all_eq_eq_used},~\ref{fig:cub_sorted_all_eq_eq_used}. These figures show how the final configuration ranking changes during planning, and that the shortest paths do not always end at the closest goal configuration to the start configuration.

\FloatBarrier
\section{Discussion}
In our experiments using more than one goal configuration resulted in the planner finding shorter paths that were quicker to execute, even when the first goal configuration was the closest to the start. In most cases using five goal configurations was enough to provide a significant improvement. Using more than 32 goal configurations often did not result in further performance increases. Using multiple goal configurations allows the planner to find paths to alternate goals that might be less obstructed by obstacles.

Our results are consistent with Dragan~et~al~\cite{Dragan2011} who found that using goal sets containing 27 configurations improved the cost of paths found by the CHOMP trajectory optimiser by 43\% on average. In our experiments we found that using 32 goal configurations resulted in the planner finding paths that were 65\% (Fig.~\ref{fig:vpr_random_all_eq_len_exec}), 15\% (Fig.~\ref{fig:vpr_sorted_all_eq_len_exec}), 55\% (Fig.~\ref{fig:cub_random_all_eq_len_exec}) and 55\% (Fig.~\ref{fig:cub_sorted_all_eq_len_exec}) shorter than those obtained using only one goal configuration at various stages of planning.

Figures~\ref{fig:vpr_sorted_all_eq_len_exec}~and~\ref{fig:cub_sorted_all_eq_len_exec} show that the planner managed to find shorter paths using a goal set compared to using only the closest goal configuration to the start. This means that sets of goal configurations should be used, even when the closest goal configuration to the start is known. 

The planner often finished with solutions that used one of the closest 20 configurations to the start in the vine pruning experiments (Figs.~\ref{fig:vpr_random_all_eq_eq_used},~\ref{fig:vpr_sorted_all_eq_eq_used}), and one of the 50 closest configurations in the cubicle picking experiments (Figs.~\ref{fig:cub_random_all_eq_eq_used},~\ref{fig:cub_sorted_all_eq_eq_used}). This is consistent with the lack of performance improvements we saw by using more than 32 goal configurations in both experiments. It could be that lower (higher number) ranked goal configurations were too far away to be useful even if they were relatively unobstructed by obstacles. 

Using a set of goal configurations may have worked well because it was likely to contain a `good' goal configuration for the query. Alternatively, it may have worked well because the extra goal configurations may have allowed the planner to quickly find high-quality intermediate solutions, reducing the region of configuration space that was sampled through its use of the informed heuristic~\cite{Gammell2014}.

It is possible that good performance could be achieved using one high quality goal configuration. Some computation time could be saved if only one goal configuration had to be found using an inverse kinematics routine. This could be achieved if an inverse kinematic solver was be biased toward solutions in particular regions of configuration space using an appropriate heuristic, although it is unclear what heuristic should be used.

In our experiments we have shown that using a set of goal configurations resulted in shorter paths than using the closest goal configuration to the start (Figs.~\ref{fig:vpr_sorted_all_eq_len_exec}~and~\ref{fig:cub_sorted_all_eq_len_exec}). This means that a heuristic used to guide the inverse kinematics solver should not only consider the proximity of the goal to the start.

\section{Conclusion}
We found that using multiple goal configurations allowed our path planner to find paths that were shorter and faster to execute. These extra goals meant that the planner was able to find paths to different goal configurations that may be shorter compared to a planner that chooses one goal configuration arbitrarily. In a grape vine pruning robot arm experiment our proposed planner reduced execution times by 58\%.

\bibliography{library}

\begin{thebibliography}{10}

\bibitem{Akgun2011}
B.~Akgun and M.~Stilman.
\newblock {Sampling heuristics for optimal motion planning in high dimensions}.
\newblock {\em International Conference on Intelligent Robots and Systems},
  2011.

\bibitem{Bac2014}
W.~Bac, E.~van Henten, J.~Hemming, and Y.~Edan.
\newblock {Harvesting Robots for High-value Crops : State-of-the-art Review and
  Challenges Ahead}.
\newblock {\em Journal of Field Robotics}, 31(6), 2014.

\bibitem{Berenson2009}
D.~Berenson and D.~Ferguson.
\newblock {Manipulation Planning with Workspace Goal Regions}.
\newblock In {\em International Conference on Robotics and Automation}, pages
  618--624, 2009.

\bibitem{Berenson2011}
D.~Berenson, S.~Srinivasa, and J.~Kuffner.
\newblock {Task Space Regions: A framework for pose-constrained manipulation
  planning}.
\newblock {\em International Journal of Robotics Research}, 30(12):1435--1460,
  2011.

\bibitem{Bertram2006}
D.~Bertram, J.~Kuffner, R.~Dillmann, and T.~Asfour.
\newblock {An integrated approach to inverse kinematics and path planning for
  redundant manipulators}.
\newblock {\em International Conference on Robotics and Automation}, pages
  1874--1879, 2006.

\bibitem{BotterillVPR}
T.~Botterill, S.~Paulin, R.~Green, S.~Williams, J.~Lin, V.~Saxton, S.~Mills,
  X.~Chen, and S.~Corbett-Davies.
\newblock {A robot system for pruning grape vines}.
\newblock {\em The Journal of Field Robotics}, 2016.

\bibitem{POMPChoudhury2016}
S.~Choudhury, C.~Dellin, and S.~Srinivasa.
\newblock {Pareto-Optimal Search over Configuration Space Beliefs for Anytime
  Motion Planning}.
\newblock In {\em International Conference on Intelligent Robots and Systems},
  2016.

\bibitem{Coleman2015}
D.~Coleman, I.~Suscan, M.~Moll, K.~Okada, and N.~Correll.
\newblock {Experience-Based Planning with Sparse Roadmap Spanners}.
\newblock In {\em International Conference on Robotics and Automation}, pages
  2--7, 2015.

\bibitem{Correll2016}
N.~Correll, K.~E. Bekris, D.~Berenson, O.~Brock, A.~Causo, K.~Hauser, K.~Okada,
  A.~Rodriguez, J.~M. Romano, and P.~R. Wurman.
\newblock {Lessons from the Amazon Picking Challenge}.
\newblock {\em Arxiv}, 6(1):1--14, 2016.

\bibitem{Dalibard2009}
S.~Dalibard, A.~Nakhaei, F.~Lamiraux, and J.-p. Laumond.
\newblock {Whole-Body Task Planning for a Humanoid Robot : a Way to Integrate
  Collision Avoidance}.
\newblock In {\em International Conference on Humanoid Robots}, pages 355--360,
  2009.

\bibitem{Dragan_2011_6887}
A.~Dragan, G.~Gordon, and S.~Srinivasa.
\newblock {Learning from Experience in Manipulation Planning: Setting the Right
  Goals}.
\newblock In {\em International Symposium on Robotics Research}, jul 2011.

\bibitem{Dragan2011}
A.~D. Dragan, N.~D. Ratliff, and S.~S. Srinivasa.
\newblock {Manipulation planning with goal sets using constrained trajectory
  optimization}.
\newblock In {\em International Conference on Robotics and Automation}, pages
  4582--4588, may 2011.

\bibitem{Drumwright2006}
E.~Drumwright and V.~Ng-thow hing.
\newblock {Toward Interactive Reaching in Static Environments for Humanoid
  Robots}.
\newblock In {\em International Conference on Intelligent Robots and Systems},
  pages 846--851, 2006.

\bibitem{Ellekilde2013}
L.-P. Ellekilde and H.~G. Petersen.
\newblock {Motion planning efficient trajectories for industrial bin-picking}.
\newblock {\em The International Journal of Robotics Research},
  32(9-10):991--1004, 2013.

\bibitem{Gammell2014}
J.~D. Gammell, S.~S. Srinivasa, and T.~D. Barfoot.
\newblock {Informed RRT*: Optimal Sampling-based Path Planning Focused via
  Direct Sampling of an Admissible Ellipsoidal Heuristic}.
\newblock In {\em International Conference on Intelligent Robots and Systems},
  2014.

\bibitem{Hawkins2013}
K.~P. Hawkins.
\newblock {Analytic Inverse Kinematics for the Universal Robots UR5/UR10 Arms},
  2013.

\bibitem{Hirano2005}
Y.~Hirano, K.-i. Kitahama, and S.~Yoshizawa.
\newblock {Image-based Object Recognition and Dexterous Hand / Arm Motion
  Planning Using RRTs for Grasping in Cluttered Scene}.
\newblock In {\em International Conference on Intelligent Robots and Systems},
  pages 2--7, 2005.

\bibitem{Jordan2013}
M.~Jordan and A.~Perez.
\newblock {Optimal Bidirectional Rapidly-Exploring Random Trees Random Trees}.
\newblock Technical report, MIT, 2013.

\bibitem{Keselman2014}
L.~Keselman, E.~Verriest, and P.~A. Vela.
\newblock {Forage RRT - An Efficient Approach To Task-Space Goal Planning for
  High Dimensional Systems}.
\newblock In {\em International Conference on Robotics and Automation}, pages
  1572--1577, 2014.

\bibitem{Klemm2015}
S.~Klemm, J.~Oberlander, A.~Hermann, A.~Roennau, and T.~Schamm.
\newblock {RRT*-Connect : Faster , Asymptotically Optimal Motion Planning}.
\newblock In {\em International Conference on Robotics and Biomimetrics}, 2015.

\bibitem{Lee2014a}
J.~J.~H. Lee, K.~Frey, R.~Fitch, and S.~Sukkareith.
\newblock {Fast Path Planning for Precision Weeding}.
\newblock In {\em Australasian Conference on Robotics and Automation}, 2014.

\bibitem{Pan2012}
J.~Pan, S.~Chitta, and D.~Manocha.
\newblock {FCL: A general purpose library for collision and proximity queries}.
\newblock {\em International Conference on Robotics and Automation}, 2012.

\bibitem{Paulin2015b}
S.~Paulin, T.~Botterill, X.~Chen, and R.~Green.
\newblock {A specialised collision detector for grape vines}.
\newblock In {\em The Australasian Conference on Robotics and Automation},
  2015.

\bibitem{Paulin2016a}
S.~Paulin, T.~Botterill, R.~Green, and X.~Chen.
\newblock {Integrating asymptotically-optimal path planning with local
  optimization}.
\newblock {\em Arxiv}, pages 1--14, 2016.

\bibitem{Phillips2012}
M.~Phillips, B.~J. Cohen, S.~Chitta, and M.~Likhachev.
\newblock {E-Graphs: Bootstrapping Planning with Experience Graphs}.
\newblock {\em Robotics Science and Systems}, 2012.

\bibitem{Ratliff2009}
N.~Ratliff, M.~Zucker, J.~A. Bagnell, and S.~Srinivasa.
\newblock {CHOMP: Gradient optimization techniques for efficient motion
  planning}.
\newblock {\em International Conference on Robotics and Automation}, 2009.

\bibitem{Stilman2007}
M.~Stilman.
\newblock {Task constrained motion planning in robot joint space}.
\newblock {\em 2007 IEEE/RSJ International Conference on Intelligent Robots and
  Systems}, pages 3074--3081, 2007.

\bibitem{Stollenga2013}
M.~Stollenga, L.~Pape, M.~Frank, and F.~Alexander.
\newblock {Task-Relevant Roadmaps : A Framework for Humanoid Motion Planning}.
\newblock In {\em International Conference on Intelligent Robots and Systems},
  pages 5772--5778, 2013.

\bibitem{VandeWeghe2007}
M.~{Vande Weghe}, D.~Ferguson, and S.~S. Srinivasa.
\newblock {Randomized Path Planning for Redundant Manipulators without Inverse
  Kinematics}.
\newblock In {\em International Conference on Humanoid Robots}, 2007.

\bibitem{Zucker2013}
M.~Zucker, N.~Ratliff, a.~D. Dragan, M.~Pivtoraiko, M.~Klingensmith, C.~M.
  Dellin, J.~a. Bagnell, and S.~S. Srinivasa.
\newblock {CHOMP: Covariant Hamiltonian optimization for motion planning}.
\newblock {\em The International Journal of Robotics Research},
  32(9-10):1164--1193, 2013.

\end{thebibliography}

\appendix
\section{Joint limits}
\label{app:joint_limiting}
\begin{figure}
	\centering
	\subfloat[UR5 in self collision.]{\includegraphics[width=0.5\linewidth]{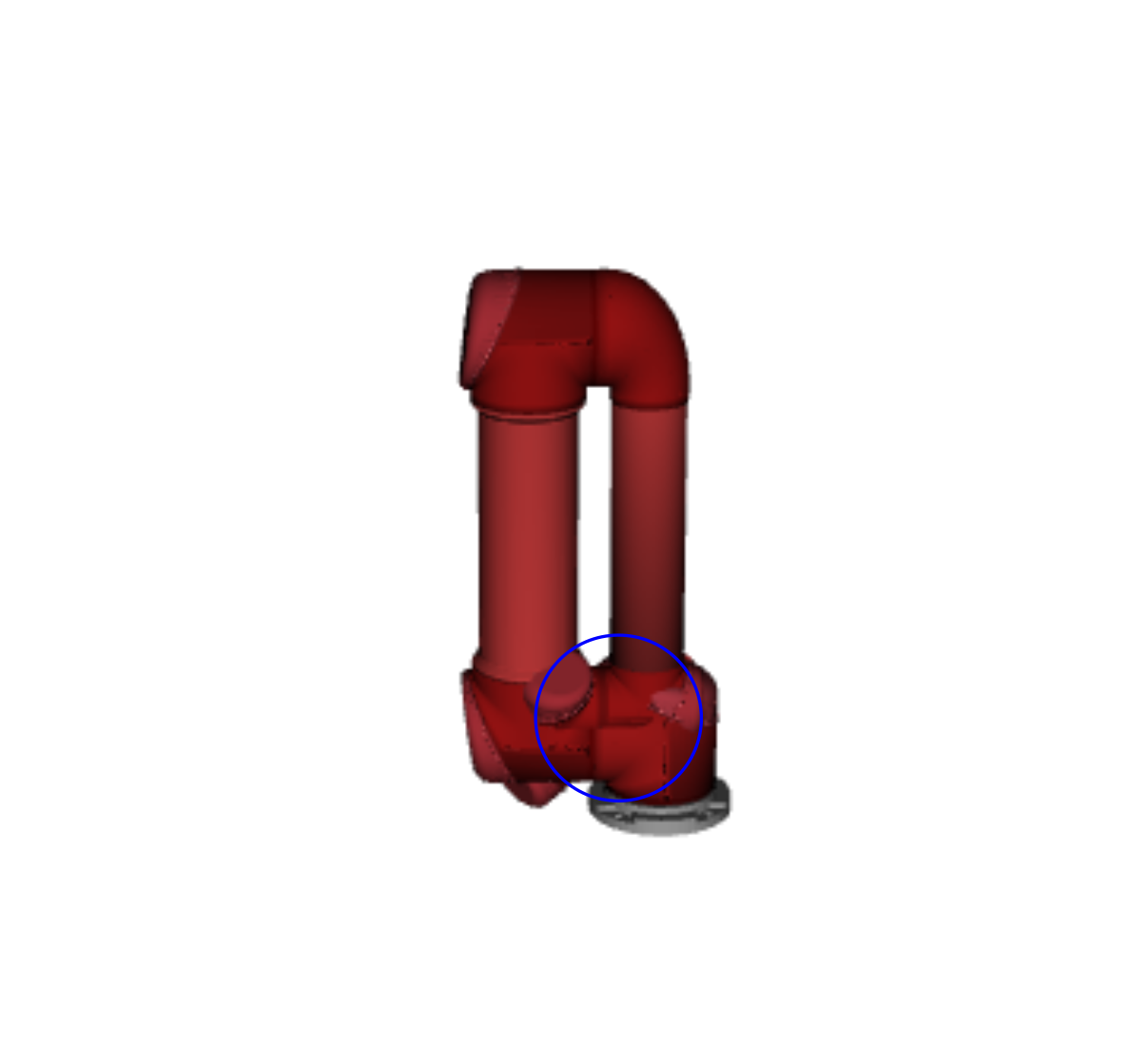}}
	
	\subfloat{\includegraphics[width=\linewidth]{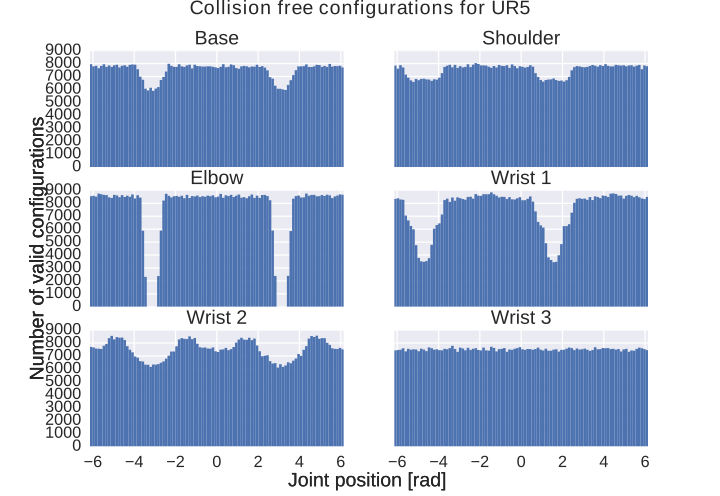}}
	\caption{The UR5 has a self collision when the elbow joint is close to $\pm \pi$, independent of the positions of the other joints. This makes the UR5's configuration space disjoint depending on the position of the elbow joint.}
	\label{fig:ur5_elbow_disjoint}
\end{figure}

The elbow joint of the UR5 robot arm used in our experiments was limited to the range $[-\pi, \pi)$. This is because the arm has a self collision when the elbow joint is close to $\pm \pi$ as shown in Fig.~\ref{fig:ur5_elbow_disjoint}. This self collision causes the UR5's configuration space to be split into three disjoint sets depending on whether the elbow joint is in $[-2\pi, -\pi)$, $[\pi, \pi)$ or $[\pi, 2\pi)$. Since these sets are disjoint, it is not possible to find a collision-free path where the start and goal positions of the elbow joint are in different sets. This has been breaking the path planning with the UR5 for some time and had been thought to be a problem with Moveit. We have reported this issue to the Universal Robot repository on GitHub. 

\begin{figure}
	\centering
	\subfloat[UR5 in collision with mounting wall.]{\includegraphics[angle=270,width=0.25\linewidth]{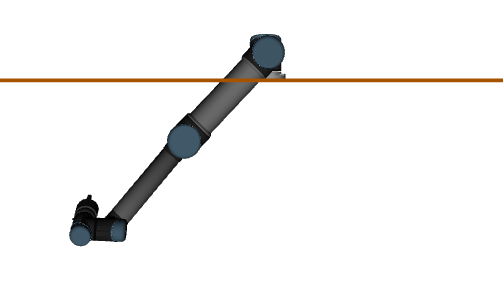}}
	
	\subfloat{\includegraphics[width=\linewidth]{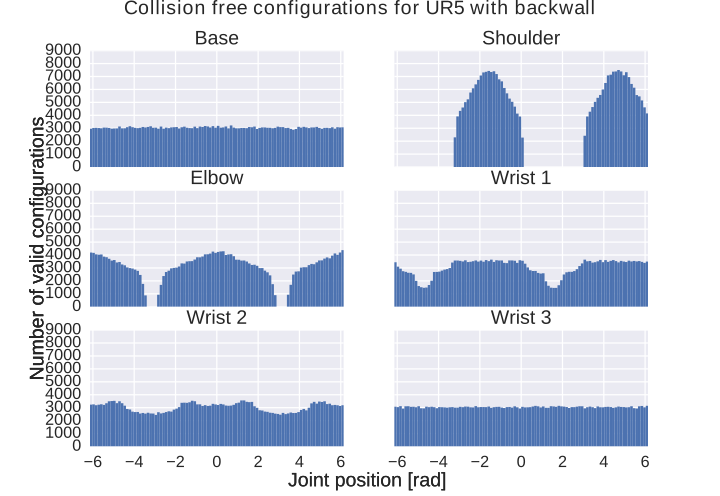}}
	\caption{Mounting the UR5 on a flat surface (brown) causes its configuration space to be disjoint depending on the position of the shoulder lift joint.}
	\label{fig:ur5_disjoint_config_space_backwall}
\end{figure}

We limited the range of the shoulder lift joint to $[-\pi, 0)$ in our experiments with the vine pruning robot. This is because the vine pruning robot has a back wall, which splits the UR5's configuration space as shown in Fig.~\ref{fig:ur5_disjoint_config_space_backwall}.

\end{document}